# Thermal deflection decoupled 6-DOF pose measurement of hexapods

Vinayak J. Kalas[1,3], Alain Vissière[2], Olivier Company[1], Sébastien Krut[1], Pierre Noiré[3], Thierry Roux[3], François Pierrot[1]

[1]*LIRMM, University of Montpellier, CNRS, 34095 Montpellier, France*
[2]*Laboratoire Commun de Métrologie (LNE-CNAM), 61 Rue du Landy, 93210 La Plaine-Saint Denis, France*
[3]*Symétrie, 10 Allée Charles Babbage, 30035 Nîmes, France*

*Email: vinayak.kalas@lirmm.fr*

**Abstract**

Calibration is crucial for hexapods with high-accuracy positioning capability. Many of these calibration procedures require measurement of hexapod's platform pose (position and orientation) at constant temperature. Consequently, thermal deflection of the hexapod's platform during pose measurements impacts the accuracy of calibrated parameters. This paper presents a method to eliminate the impact of thermal deflection of hexapods on their pose measurement. In this method, a reference pose is measured before each measurement of any particular pose. The measurements of reference pose are used to estimate thermal deflection of hexapod's legs. This is in turn used to estimate and correct the resulting pose error at the particular pose to be measured. The advantage of using the proposed method is demonstrated experimentally by means of pose measurements of a high-precision hexapod using a CMM.

Keywords: pose measurement, accuracy, thermal error, hexapod, positioning system

## 1. Introduction

Hexapods are increasingly being used for high-accuracy 6-DOF positioning applications. Different types of calibrations (geometric, elastostatic, etc.) are performed to achieve their high-acccuracy positioning capability. These calibrations are sensitive to the accuracy of platform pose measurements. Many of these calibration procedures require the hexapod to be at a constant temperature during pose measurements using the conventional method. When this condition is violated, the accuracy of calibrated parameters is affected.

This paper presents a method to solve the aforementioned problem. The presented method eliminates the effects of thermal deflection of the hexapod on platform pose measurement. The advantage of the presented method has been experimentally demonstrated by means of pose measurements of a high-precision hexapod performed using a CMM. This paper is organised as follows: section 2 outlines the conventional pose measurement method and its drawback. Section 3 presents the thermal deflection decoupled pose measurement method for hexapods. Section 4 presents the experimental study followed by conclusions in section 5.

## 2. Conventional pose measurement method and its drawback

Pose measurements are always made by measuring points using a measurement system which has a coordinate frame $M$ attached to it. All the points are measured with respect to this coordinate frame. The requirement is to measure the coordinate frame fixed to the platform (platform frame), $S_i$, when the hexapod is in any $i^{th}$ arbitrary configuration with respect to another coordinate frame $O$. The frame $O$ is defined at some geometric landmark on the hexapod.

The conventional method to measure the 6-DOF pose vector of the platform frame of an arbitrary pose $S_1$ with respect to frame $O$ is illustrated in figure 1. The frame $O$ is measured with respect to $M$ first. This coordinate frame is measured with the legs having a temperature set $t_1 = [t_{11}, t_{12}, .., t_{16}]$, where $t_{1i}$ is the temperature of the $i^{th}$ leg during this measurement. Let this measured frame be called $O^{t_1}$. Any coordinate frame can be measured with respect to $M$ using different methods and depends on the measurement setup available [1-4]. After measuring $O^{t_1}$, the platform frame $S_1$ is measured. This measurement happens with the legs having a temperature set $t_2 = [t_{21}, t_{22}, .., t_{26}]$. Let this coordinate frame be called $S_1^{t_2}$. The transformation between the coordinate frames $O^{t_1}$ and $S_1^{t_2}$, written as $T(O^{t_1}, S_1^{t_2})$, is then computed. Consequently, the corresponding 6-DOF pose vector $X_{S_1^{t_2}}^{O^{t_1}}$ is obtained.

From the description presented above, it can be easily seen that the measured transformation would have been different if the hexapod's legs would have had the temperature set $t_1$. The platform frame in this case ($S_1^{t_1}$) would have a different pose vector $X_{S_1^{t_1}}^{O^{t_1}}$. This is due to the thermal deflection of the legs of the hexapod with the change in their temperatures from set $t_1$ to set $t_2$. Temperature change also affects other dimensions of the hexapod. However, for most hexapods, the thermal deflection of legs is much higher than that of the other parts because: (a) the legs generally have larger dimensions (length) as compared to the other parts, and (b) driving motors are mounted on/near the legs which heat the legs more than the other parts.

To understand the problem with the conventional pose measurement method, consider the case of geometric calibration. When the conventional pose measurement method is used in geometric calibration, different platform poses are measured with the legs at different temperatures. This can happen due to heating supplied by motors or the surrounding air. Consequently, different measured poses have the influence of different magnitudes of thermal deflections of legs. This situation is problematic for geometric calibration. This is because, in geometric calibration, the difference between measured and commanded poses of the platform are considered to be a consequence of errors in known geometric parameters only [5]. Hence, the accuracy of calibrated geometric parameters is adversely affected when the conventional pose measurement method is used. The same problem can also impact accuracy of identified stiffness parameters in hexapod elastostatic calibration [6].

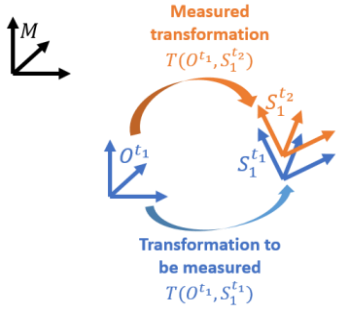

**Figure 1.** Illustration of the conventional pose measurement method to measure the coordinate frame of an arbitrary pose $S_1$ with respect to a coordinate frame $O$

## 3. Thermal deflection decoupled pose measurement method

Figure 2 illustrates the proposed method to measure the 6-DOF pose vector of the platform frame of an arbitrary pose $S_1$ with respect to frame $O$. In this, frame $O^{t_1}$ is measured first (with the legs having temperature set $t_1$). Immediately after this, the platform is moved to a reference pose $R$. This pose is measured quickly such that the measurement happens with the legs having the temperature set $t_1$. Let this measured frame be called $R^{t_1}$. The platform can then be moved to any arbitrary pose (frame $S_1$) and the platform frame can be measured with the legs having a temperature set $t_2$. This measured frame is $S_1^{t_2}$. An additional measurement of the frame $R$ must be performed quickly before/after measuring $S_1^{t_2}$. This measurement must be carried out with the hexapod's legs having the temperature set $t_2$ (measured coordinate frame: $R^{t_2}$).

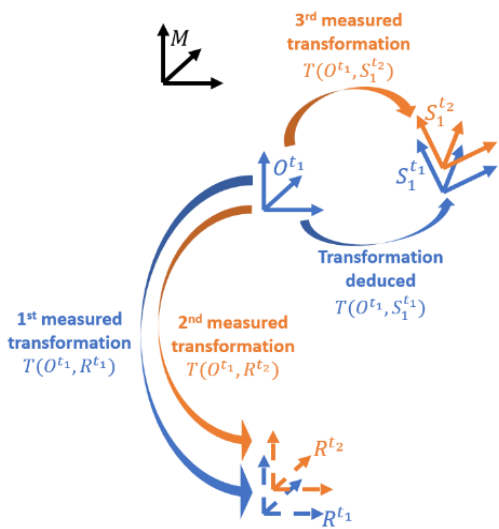

**Figure 2.** Illustration of the thermal deflection decoupled pose measurement method to measure the coordinate frame of an arbitrary pose $S_1$ with respect to a coordinate frame $O$

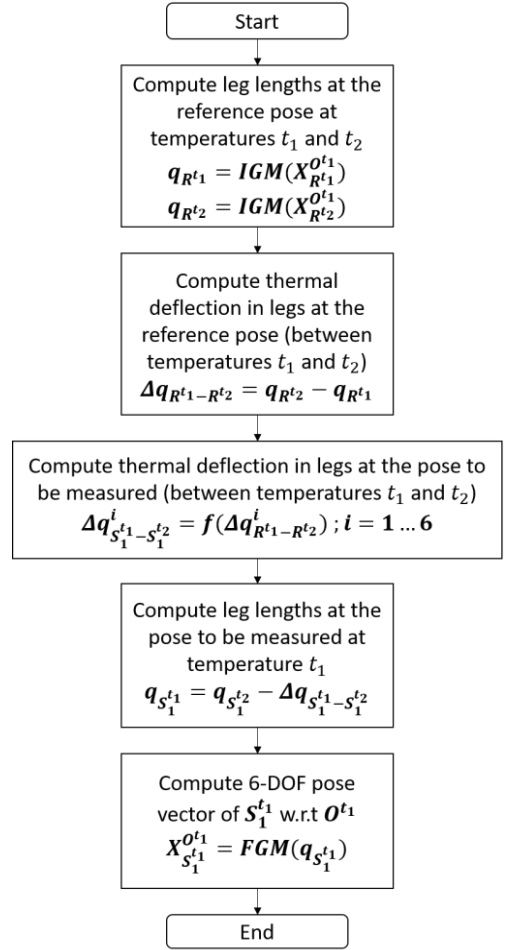

**Figure 3.** Flowchart of procedure to post-process the measured data to obtain the required pose vector in the thermal deflection decoupled pose measurement method

Figure 3 illustrates the method to obtain the necessary pose vector $X_{S_1^{t_1}}^{O^{t_1}}$ using the measurement data obtained from the measurements outlined above. The measurement procedure described can be used to obtain three transformations: $T(O^{t_1}, R^{t_1})$, $T(O^{t_1}, R^{t_2})$ and $T(O^{t_1}, S_1^{t_2})$. Consequently, the corresponding 6-DOF pose vectors, $X_{R^{t_1}}^{O^{t_1}}$, $X_{R^{t_2}}^{O^{t_1}}$ and $X_{S_1^{t_2}}^{O^{t_1}}$, can be obtained. These pose vectors can be used to get the corresponding leg lengths of the hexapod by using the inverse geometric model ($IGM$) of the hexapod [7]. $q_{R^{t_1}}$, $q_{R^{t_2}}$ and $q_{S_1^{t_2}}$ are the arrays containing the leg lengths of the hexapod corresponding to pose vectors $X_{R^{t_1}}^{O^{t_1}}$, $X_{R^{t_2}}^{O^{t_1}}$ and $X_{S_1^{t_2}}^{O^{t_1}}$, respectively. $q_{R^{t_1}}$ and $q_{R^{t_2}}$ can then be used to compute the thermal deflection the hexapod's legs corresponding to temperature change from set $t_1$ to set $t_2$, with the platform at pose $R$. Let the array containing these leg deflections be called $\Delta q_{R^{t_1}-R^{t_2}}$ and let $\Delta q_{R^{t_1}-R^{t_2}}^i$ be the deflection of the $i^{th}$ leg. The thermal deflection due to temperature change of legs from set $t_1$ to set $t_2$ of the $i^{th}$ leg of the hexapod at the arbitrary pose $S_1$, $\Delta q_{S_1^{t_1}-S_1^{t_2}}^i$, can then be estimated easily. The task here is to find the thermal deflection of the legs with lengths $q_{S_1^{t_2}}$, corresponding to temperature change of legs from set $t_1$ to set $t_2$, when the thermal deflection of the same legs with lengths $q_{R^{t_1}}$ are known. The method to perform this computation must respect the dimensions and material properties of the components of the leg assembly. $\Delta q_{S_1^{t_1}-S_1^{t_2}}$ can then be subtracted from $q_{S_1^{t_2}}$ to obtain $q_{S_1^{t_1}}$. $q_{S_1^{t_1}}$ is the array containing the leg lengths when the platform is at the arbitrary pose $S_1$ and the legs have temperature set $t_1$. Finally, the necessary pose vector $X_{S_1^{t_1}}^{O^{t_1}}$ can be obtained by using forward geometric model ($FGM$) of the hexapod [7] corresponding to $q_{S_1^{t_1}}$. When multiple platform poses shall be measured using this method while leg temperatures change, the measured poses will not have the influence of different magnitudes of thermal deflections of legs. Hence, the drawback of the conventional method can be overcome using this method.

## 4. Experiments and results

This section presents the details of an experimental study performed to compare the conventional and proposed methods. The legs of the hexapod were heated during this experiment to control and slightly exaggerate heating in legs. This was done to clearly show the advantage of the proposed pose measurement method over the conventional method. Figure 4 shows the test setup used for this experimental study. A high-precision positioning hexapod from Symétrie [8] was used for this study and this has a repeatability of ±0.5μm in translations and ±2.5μrad in rotations. A flexible electric heating mat was fixed to each leg to facilitate heating. Thermocouples were used to measure the temperature of each leg and the surrounding air. Precision balls were fixed to the hexapod's platform which were used for measuring the coordinate frame fixed to the platform. The measurements were performed using a LK-METRIS CMM equipped with a RENISHAW SP25 touch probe. The uncertainty of points measured using this CMM, quantified using the MPE$_P$ value [9], is about ±2 μm.

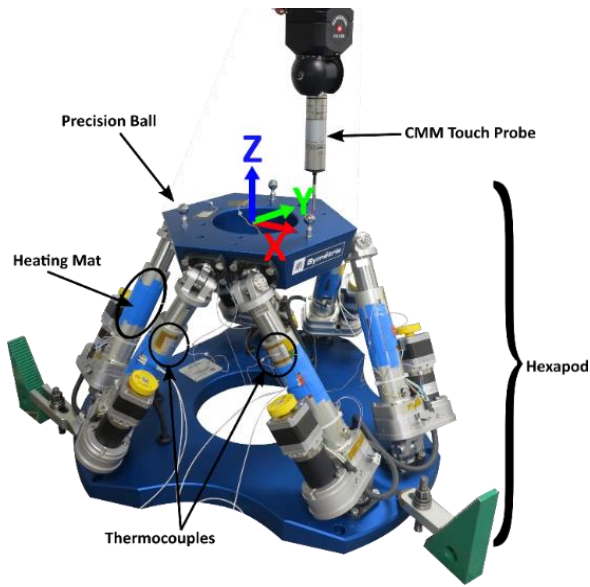

**Figure 4.** Test setup

Poses of the platform of the hexapod used in this study are defined by a coordinate frame fixed to the platform at its center. The position and orientation of this platform frame is pre-defined using holes and planes which are precisely machined on the platform in the manufacturing phase. The coordinate frame with respect to which any pose of the platform is defined ($O$) is the platform frame with the hexapod in a certain configuration. In this configuration, the hexapod's actuators/legs are locked at a certain length and all the legs have the same length. This pose will be referred to as *zero pose*. The pose vector of the platform pose is written such that the first three elements represent the translations and last three represent rotations, along and about the X, Y and Z axes of $O$, respectively. The hexapod's platform has the pose vector $[0\ mm\ 0\ mm\ 0\ mm\ 0°\ 0°\ 0°]$ when the platform is in zero pose (figure 5).

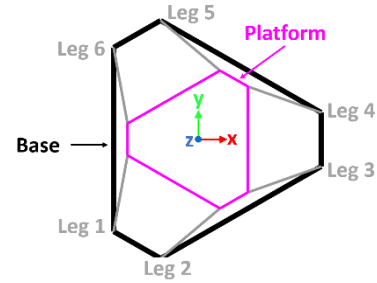

**Figure 5.** Hexapod with platform in $[0mm\ 0mm\ 0mm\ 0°\ 0°\ 0°]$ pose (top view)

In order to perform platform pose measurements, the platform frames must be measured with respect to the coordinate frame of the CMM. In order to do this, the platform frame is first measured by measuring the holes and planes precisely machined on the platform, with the hexapod in any configuration. Quickly after this, the centers of the three precision balls are identified by measuring several points on the surface of each ball. Consequently, the position and orientation of the platform frame is known with respect to the three balls. Note that this relationship is fixed since the balls and the platform frame are fixed to the platform. Now, to measure the platform frame with the platform in any pose, just the centers of the three balls need to be measured. The corresponding platform frame can then be identified using the relationship known between the platform frame and the centers of balls.

In this experimental study, the pose to be measured (refered to as *measurement pose* from here), called $S_1$ in sections 2 and 3, was the zero pose. This pose was chosen to be measured in order to facilitate the ease of understanding results as the hexapod is symmetrical in this configuration (see figure 5). The reference pose $R$ to be used in the proposed pose measurement method had the pose vector $[0\ mm\ 0\ mm\ -40\ mm\ 0°\ 0°\ 0°]$. The necessary measurements were made to perform the pose measurement as per the proposed method (see section 3). Note that (a part of) these measurements can also be used for performing pose measurements as per the conventional method. 10 trials of measurements were performed and the hexapod's legs were heated during this using the electric heating mats. The measurements were then post-processed as per the conventional and proposed methods.

In the proposed pose measurement method, the thermal expansion of the legs with the platform in measurement pose had to be predicted. This had to be done using the measured thermal expansions of the legs at the reference pose (see figure 3). The following logic was used for this: the legs of the hexapod used in this study could be divided length-wise into an Aluminium part of fixed length and a Steel part of variable length. When the platform is moved from one pose to another, the Steel parts of legs change their lengths to achieve the new required lengths. When the thermal expansion of legs at the reference pose were measured, the corresponding thermal expansions of the Aluminium and the Steel parts could be determined. This could be done because the lengths and the thermal expansion coefficients of the two parts were known. The length of each leg and the corresponding length of the Steel part, with the hexapod in the measurement pose, were also known. The thermal expansion of the Steel part of each leg measured in reference pose was then appropriately scaled to estimate the thermal expansion of the Steel part of each leg in measurement pose. The thermal expansion of the Aluminium part was same for the reference and measurement poses as this part does not change its length. The total thermal expansion of each leg at the measurement pose was then obtained by adding the corresponding thermal expansions of the Steel and Aluminium parts.

Figure 6 shows the pose parameters of the measurement pose measured by using the conventional and proposed methods. $Tx_{mes}$, $Ty_{mes}$ and $Tz_{mes}$ are the components of measured pose vector corresponding to translations along X, Y and Z axes of the hexapod, respectively. $Rx_{mes}$, $Ry_{mes}$ and $Rz_{mes}$ are the components of measured pose vector corresponding to rotations about X, Y and Z axes of the hexapod, respectively. Figure 7 shows the temperatures measured at different locations during this test.

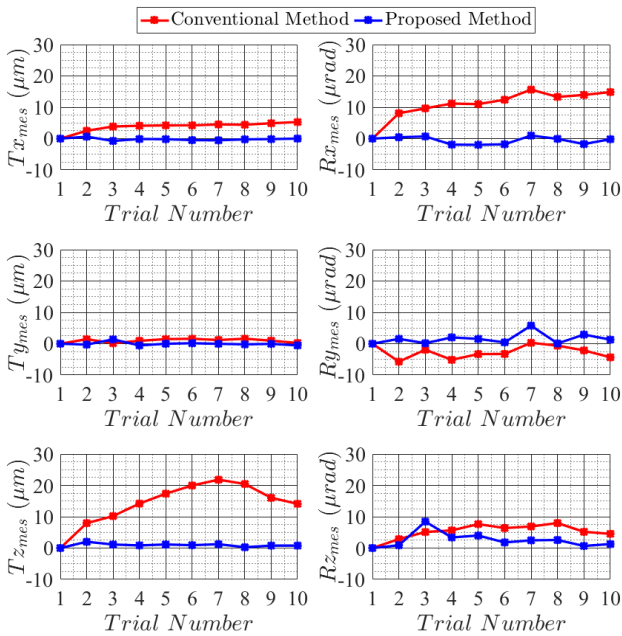

**Figure 6.** Measured pose parameters using conventional and proposed methods with the platform in zero pose

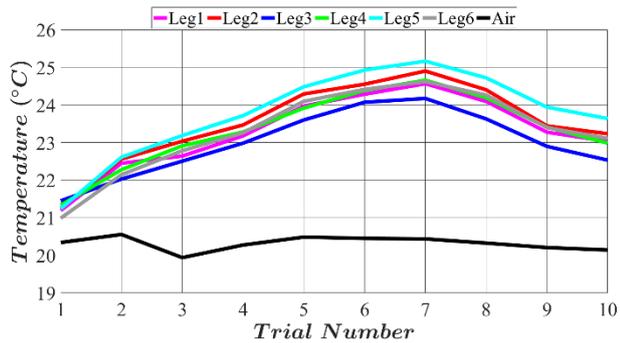

**Figure 7.** Measured temperatures

Figure 6 shows that the measured pose parameters obtained using the conventional method deviates significantly with every trial. These observed deviations can be correlated with the change in temperature between trials. The trend of deviation of $Tz_{mes}$ using the conventional method is similar to the trend of the change in temperature of all legs. This behaviour is logical given the orientation of all legs in the considered measurement pose. Also, deviation seen in $Tx_{mes}$ using conventional method increases with every consecutive trial until the end. This can be explained by the temperatures measured in legs 2, 3, 4 and 5. The temperatures of legs 2 and 5 are higher than those of legs 3 and 4 during the test and this difference increases with every consecutive trial until the end. Consequently, legs 2 and 5 push the platform more in positive X-direction as compared to legs 3 and 4 pushing it in the opposite direction. Furthermore, deviation seen in $Rx_{mes}$ using conventional method also increases with every consecutive trial until the end. This can be explained by the difference in temperatures of legs 3 and 5 (with leg 5 heating more than leg 3) which follows a similar trend. Consequently, leg 5 pushes the platform more about the X-axis as compared to leg 3 and results in a positive rotational deviation about the X-axis with every consecutive trial. The pose parameters measured using the proposed method do not deviate with change in temperature of hexapod's legs, unlike the ones measured using conventional method. It is, therefore, clear that the proposed method is effective in eliminating the influence of thermal deflection of the hexapod on the measured pose parameters.

## 5. Conclusion

This paper presented a method to eliminate the influence of thermal deflection of a hexapod on the measured 6-DOF pose of its platform. The presented method is validated experimentally by means of pose measurements of a high-precision hexapod using a CMM. Results of this experimental study confirm the efficacy of the proposed method. In future work, the use of measured temperature data to eliminate the influence of hexapod's thermal deflection on the measured 6-DOF pose of its platform will be developed.

**Acknowledgements**

This work was supported by Agence Nationale de la Recherche (ANR), France (Project ANR-15-LCV3-0005). The authors hereby express their gratitude for the support.

**References**

[1] Zhang G, Du J and To S 2014 Calibration of a small size hexapod machine tool using coordinate measuring machine *Proceedings of the Institution of Mechanical Engineers Part E J. of Process Mechanical Engineering* **230(3)** 183-197

[2] Driels M R, Swayze W and Potter S 1993 Full-pose calibration of a robot manipulator using a coordinate measuring machine *The Int. J. of Advanced Manufacturing Technology* **8(1)** 34-41

[3] Meng G, Tiemin L and Wensheng Y 2003 Calibration method and experiment of Stewart platform using a laser tracker *SMC'03 Conference Proceedings 2003 IEEE Int. Conf. on Systems, Man and Cybernetics* **3** 2797-2802

[4] Zhuang H, Masory O and Yan J 1995 Kinematic calibration of a Stewart platform using pose measurements obtained by a single theodolite *Proceedings of IEEE/RSJ Int. Conf. on Intelligent Robots and Systems* **2** 329-334

[5] Siciliano B and Khatib O 2008 *Springer handbook of robotics*, Springer

[6] Kalas VJ, Vissière A, Roux T, Company O, Krut S and Pierrot F 2018 A new efficient stiffness evaluation method to improve accuracy of hexapods *ASME 2018 International Design Engineering Technical Conferences and Computers and Information in Engineering Conference* **5A** V05AT07A045

[7] Dombre E and Khalil W 2013 *Robot manipulators: modeling, performance analysis and control*, John Wiley & sons

[8] Symétrie *Positioning hexapods* Available at: https://www.symetrie.fr/en/products/positioning-hexapods/ (Accessed: 20 December 2019)

[9] International Standards Organization 2000 *ISO 10360-1 : 2000 Geometrical product specifications (GPS) – acceptance and reverification tests for coordinate measuring machines (CMM), Part 1: Vocabulary*